\documentclass[11pt]{article}
\usepackage[final,hyperref]{acl}
\usepackage{times}
\usepackage{latexsym}
\usepackage[T1]{fontenc}
\usepackage[utf8]{inputenc}
\usepackage{microtype}
\usepackage{booktabs}
\usepackage{graphicx}
\usepackage{amsmath}
\usepackage{xcolor}
\usepackage{colortbl}
\usepackage{enumitem}
\usepackage{tikz}
\usetikzlibrary{arrows.meta, positioning, shapes.geometric}
\usepackage{fontawesome5}
\usepackage[hang,flushmargin]{footmisc}
\definecolor{lightgray}{gray}{0.93}
\definecolor{accent}{HTML}{2563EB}

\title{BlasBench: An Open Benchmark for Irish Speech Recognition \\[4pt]
{\large\normalfont\faGithub~\href{https://github.com/jyoutir/blasbench}{\texttt{github.com/jyoutir/blasbench}}}}

\author{
Jyoutir Raj\thanks{\ Corresponding author.} \\
Independent Researcher \\
\texttt{jyoutirraj@gmail.com}
\And
John Conway \\
Independent Researcher \\
\texttt{john@blasapp.com}
}

\date{}

\begin{document}
\setlength{\parindent}{0pt}
\setlength{\parskip}{0.5\baselineskip}
\maketitle

\begin{abstract}
Existing multilingual benchmarks include Irish among dozens of languages but apply no Irish-aware text normalisation, leaving reliable and reproducible ASR comparison impossible. We introduce \textbf{BlasBench}, an open evaluation harness that provides a standalone Irish-aware normaliser preserving fadas, lenition, and eclipsis; a reproducible scoring harness and per-utterance predictions released for all evaluated runs. We pilot this by benchmarking 12~systems across four architecture families on Common Voice~ga-IE and FLEURS~ga-IE. All Whisper variants exceed 100\%~WER through insertion-driven hallucination. Microsoft Azure reaches 22.2\%~WER on Common Voice and 57.5\% on FLEURS; the best open model, Omnilingual ASR~7B, reaches 30.65\% and 39.09\% respectively. Models fine-tuned on Common Voice degrade 33--43~points moving to FLEURS, while massively multilingual models degrade only 7--10---a generalisation gap that single-dataset evaluation misses.
\end{abstract}

\section{Introduction}

State-of-the-art multilingual systems fail on out-of-distribution audio in Irish. Microsoft Azure reaches 22.2\%~WER on read speech (Common Voice ga-IE) and 57.5\%~WER on the more naturalistic FLEURS ga-IE. Every Whisper variant we evaluate \citep{radford2023whisper} produces more tokens than the reference contains, exceeding 100\%~WER on both datasets (Tables~\ref{tab:cv},~\ref{tab:fleurs}). Irish presents a genuine difficulty: initial consonant mutations, eclipsis, and a spelling system that encodes grammatical contrasts mean that language-agnostic normalisation silently discards linguistically meaningful variation.

The problem is not that nobody has worked on Irish ASR. The ABAIR group at Trinity College Dublin has produced a sustained research programme culminating in their Fotheidil system at 10.9\%~WER \citep{lonergan2025fotheidil}. \citet{qian2024whisper} fine-tune Whisper on Irish. Community wav2vec2 fine-tunes exist on HuggingFace \citep{faste2022wav2vec2}. The IWSLT shared task has produced Irish--English speech translation systems \citep{ahmad2024iwslt, moslem2024synthetic}. Multilingual benchmarks---FLEURS \citep{conneau2022fleurs}, MMS \citep{pratap2023mms}, ML-SUPERB \citep{shi2023mlsuperb}---include Irish among dozens or hundreds of languages. These efforts span architecture families and data regimes, yet they have not been placed on a common footing.

The gap is methodological rather than empirical. ABAIR's best result is on private M\'{i}leGl\'{o}r data; their published Common Voice and FLEURS-R figures cover only their own systems. Community models self-report on different Common Voice versions with different normalisers. Multilingual benchmarks include Irish without Irish-specific normalisation or targeted analysis.\footnote{ML-SUPERB and XTREME-S \citep{conneau2022xtremes} include Irish data points but do not compare end-user ASR systems on Irish specifically.} We did not identify a prior open, Irish-specific comparison of end-user ASR systems across architecture families under a shared Irish-aware evaluation protocol.

We present BlasBench,\footnote{Harness: \href{https://github.com/jyoutir/blasbench}{github.com/jyoutir/blasbench}. Per-utterance predictions and aggregate results for every run in this paper are released as \href{https://github.com/jyoutir/blasbench/releases/tag/v0.1.0}{v0.1.0}.} an open benchmark and evaluation harness for Irish ASR. The harness applies Irish-aware text normalisation before scoring, so that orthographic variants reflecting mutation and dialect do not inflate error rates artificially. A 12-model comparison across commercial, large-scale multilingual, and fine-tuned systems constitutes the first use of the benchmark; reproducible, language-aware scoring is also a prerequisite for any automated research pipeline that requires a stable signal. Dedicated low-resource benchmarks for Urdu \citep{arif2024wer} and Scottish Gaelic \citep{klejch2025gaelic} have made the same case for their respective languages; BlasBench brings it to Irish.

\section{The BlasBench Benchmark}

\subsection{Task and scope}

BlasBench evaluates Irish speech-to-text systems. Given Irish audio, a system produces Irish text; the benchmark scores the output against a reference transcription using WER and CER, computed with Irish-aware normalisation.

\subsection{Datasets}

Two public datasets, chosen to measure both in-distribution and out-of-distribution performance:

\begin{itemize}[leftmargin=*, itemsep=2pt]
    \item \textbf{Common Voice 25.0 ga-IE} \citep{ardila2020commonvoice}: 874~community-recorded read-speech test utterances in the released BlasBench evaluation slice; the utterance count is recorded in the released run metadata. Most wav2vec2 fine-tunes we evaluate were trained on earlier CV versions.
    \item \textbf{FLEURS ga-IE} \citep{conneau2022fleurs}: 842~test utterances in the released BlasBench evaluation slice; the utterance count is recorded in the released run metadata. The speech is professionally recorded native-speaker read speech. Based on public model cards, no model in our benchmark reports FLEURS in its training mix.
\end{itemize}

The released prediction files and run metadata make each evaluation slice auditable. Using two datasets with different recording conditions is central to the benchmark design: it exposes generalisation failures invisible to single-dataset evaluation (\S\ref{sec:generalization}).

\subsection{Evaluation pipeline}

\begin{figure}[t]
\centering
\begin{tikzpicture}[
    node distance=0.4cm,
    box/.style={draw, rounded corners=3pt, fill=white, font=\small\sffamily, minimum height=0.6cm, minimum width=4.5cm, align=center, line width=0.6pt},
    accent/.style={box, fill=accent!8, draw=accent!60},
    arr/.style={-{Stealth[length=5pt]}, line width=0.6pt, color=black!50},
]
\node[box] (audio) {Audio input {\scriptsize(16kHz)}};
\node[box, below=of audio] (model) {Model wrapper {\scriptsize(any architecture)}};
\node[accent, below=of model] (norm) {Irish-aware normaliser};
\node[box, below=of norm] (score) {Global WER / CER};
\node[box, below=of score] (ci) {Bootstrap 95\% CI};

\draw[arr] (audio) -- (model);
\draw[arr] (model) -- node[right, font=\scriptsize\itshape, text=black!50, xshift=2pt] {raw text} (norm);
\draw[arr] (norm) -- (score);
\draw[arr] (score) -- (ci);
\end{tikzpicture}
\caption{BlasBench evaluation pipeline. The Irish normaliser (highlighted) is the language-specific component.}
\label{fig:pipeline}
\end{figure}
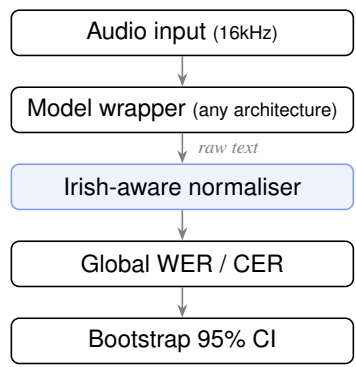

The pipeline (Figure~\ref{fig:pipeline}) proceeds in four stages:

\begin{enumerate}[leftmargin=*, itemsep=2pt]
    \item \textbf{Inference.} Audio is passed through a model wrapper that returns raw text. The wrapper API is standardised: any system that accepts 16kHz audio and returns a string can be evaluated.
    \item \textbf{Normalisation.} Both reference and hypothesis are normalised using an Irish-aware pipeline (\S\ref{sec:normaliser}).
    \item \textbf{Scoring.} WER and CER are computed via global aggregation: total substitutions, insertions, and deletions across all utterances divided by total reference units. This corpus-level approach avoids the bias of per-sentence averaging and matches the convention used in low-resource benchmarking work \citep{arif2024wer}.
    \item \textbf{Confidence intervals.} Bootstrap 95\%~CIs \citep{bisani2004bootstrap} are computed with 1{,}000~resamples at a fixed seed (42), resampling utterance-level error counts and recomputing the global aggregate. CIs for every reported point estimate are released alongside the predictions in each run's \texttt{results.json}; tables show point estimates for compactness.
\end{enumerate}

\subsection{Irish-aware normalisation}
\label{sec:normaliser}

Standard ASR normalisers lowercase text, strip punctuation, and collapse whitespace. Recent work on multilingual ASR evaluation shows that language-agnostic normalisation can materially distort reported error rates \citep{manohar2024normalization}. For Irish, it destroys linguistically significant information.

Irish uses acute accents (fadas) on five vowels: \'{a},~\'{e},~\'{i},~\'{o},~\'{u}. These are phonemically contrastive: \emph{fear} ``man'' vs \emph{f\'{e}ar} ``grass.'' Standard NFKD normalisation decomposes accented characters and may strip the combining accent. Our normaliser applies NFC first, guaranteeing fadas survive.

Irish also marks grammar through initial mutations. Lenition: \emph{bean} ``woman'' $\to$ \emph{bhean} ``his wife.'' Eclipsis: \emph{bean} $\to$ \emph{mbean} ``of women.'' Our normaliser preserves these.

ABAIR's pipeline includes text-processing components used inside Fotheidil \citep{lonergan2025fotheidil} but no Irish ASR normaliser has been released standalone. To our knowledge, BlasBench ships the first standalone open-source Irish ASR evaluation normaliser released specifically for Irish ASR scoring; unreleased internal tools may exist.

\subsection{Released artifacts}

For each (model, dataset) run, BlasBench produces:
\begin{itemize}[leftmargin=*, nosep]
    \item Per-utterance predictions with sample ID, reference, hypothesis, WER, CER
    \item Aggregate results with S/I/D breakdown and bootstrap CIs
    \item Run metadata recording dataset name/split/count, model identity, evaluation configuration, and software versions
\end{itemize}

The evaluation harness and released predictions are available at the repository linked under the title. Future systems can be compared against these predictions without re-running prior models.

\subsection{Reproducibility and extensibility}

Adding a new model requires implementing a single function: given 16kHz audio, return a string. The harness handles normalisation, scoring, CI computation, and artifact generation.

\section{Systems Evaluated}

We evaluate 12~systems spanning four architecture families.

\textbf{Whisper} \citep{radford2023whisper}: Encoder-decoder transformer. Irish is absent from Whisper's language list across v1, v2, and v3,\footnote{\texttt{ga} is absent from the \texttt{LANGUAGES} dictionary in \texttt{whisper/tokenizer.py} for v1/v2/v3, and the v3 model card provides no per-language training-data disclosure that would indicate Irish coverage.} so results here are effectively zero-shot. We test medium~(769M), large-v2~(1.5B), large-v3~(1.5B), and large-v3-turbo~(809M).

\textbf{wav2vec2 CTC}: Self-supervised XLS-R or XLSR-53 encoders with CTC heads, fine-tuned on Irish Common Voice by community contributors (315M--1B).

\textbf{Meta multilingual}: MMS-1B-All \citep{pratap2023mms} (1{,}107~languages); Omnilingual ASR (omniASR) LLM \citep{keren2025omniasr}, wav2vec2 encoders with LLM-inspired Transformer decoders covering 1{,}600+~languages, released in 300M, 1B, 3B, and 7B sizes; we test the 300M and 7B endpoints.

\textbf{Commercial}: Microsoft Azure Speech Services (ga-IE locale).

Open-weights models ran on a single NVIDIA H100 80GB SXM via RunPod ($\sim$6~GPU hours, $\sim$\$25). Azure ran via API ($\sim$\$5 at April 2026 pricing). Whisper, wav2vec2, MMS, and Azure runs all used the released harness; the omniASR runs used a separate fairseq2 driver at the time of writing, with a harness adapter pending.

\section{Results}

Tables~\ref{tab:cv} and~\ref{tab:fleurs} present WER, CER, and per-error-type breakdown on both datasets.

\begin{table*}[t!]
\centering
\small
\caption{Common Voice ga-IE (874 utterances). ABAIR is self-reported \citep{lonergan2025fotheidil} using a different normaliser. Whisper results are zero-shot. Type column: \emph{open} = weights and training data public; \emph{open-w} = weights public, training data not fully disclosed; \emph{API} = commercial cloud API; \emph{closed} = weights unavailable.}
\label{tab:cv}
\setlength{\tabcolsep}{4pt}
\begin{tabular}{clllrrrrr}
\toprule
\textsc{\#} & \textsc{model} & \textsc{type} & \textsc{arch} & \textsc{wer}$\downarrow$ & \textsc{sub} & \textsc{ins} & \textsc{del} & \textsc{cer}$\downarrow$ \\
\midrule
\emph{ref} & \emph{ABAIR / Fotheidil} & \emph{closed} & \emph{TDNN-HMM} & \emph{19.6*} & --- & --- & --- & --- \\
\midrule
1 & azure / speech-ga-IE & API & proprietary & \textbf{22.2} & 15.8 & 1.7 & 4.8 & \textbf{11.4} \\
2 & omniASR LLM 7B & open-w & w2v2+Trans & 30.6 & 25.0 & 2.5 & 3.2 & 14.6 \\
3 & Aditya3107 / xls-r-1b & open & w2v2 CTC & 32.4 & 26.4 & 1.7 & 4.3 & 12.8 \\
4 & omniASR LLM 300M & open-w & w2v2+Trans & 37.6 & 29.3 & 3.4 & 4.9 & 19.2 \\
5 & kingabzpro / xls-r-1b & open & w2v2 CTC & 45.8 & 38.2 & 3.6 & 4.0 & 18.9 \\
6 & jimregan / xlsr-53 & open & w2v2 CTC & 48.9 & 40.5 & 4.2 & 4.2 & 20.3 \\
7 & cpierse / xlsr-53 & open & w2v2 CTC & 49.4 & 41.5 & 3.9 & 4.0 & 21.0 \\
8 & mms-1b-all & open & w2v2 CTC & 54.2 & 44.1 & 2.8 & 7.4 & 21.2 \\
\midrule
9 & whisper-large-v2 & open & enc-dec & 106.0 & 73.6 & 19.9 & 12.4 & 68.5 \\
10 & whisper-large-v3 & open & enc-dec & 125.6 & 78.8 & 33.1 & 13.7 & 85.6 \\
11 & whisper-medium & open & enc-dec & 129.3 & 76.2 & 40.6 & 12.5 & 91.8 \\
12 & whisper-large-v3-turbo & open & enc-dec & 225.6 & 83.1 & 128.8 & 13.7 & 159.7 \\
\bottomrule
\end{tabular}
\end{table*}

\begin{table*}[t!]
\centering
\small
\caption{FLEURS ga-IE (842 utterances). ABAIR's 44.5\% is on FLEURS-R \citep{ma2024fleursr}, not this test set.}
\label{tab:fleurs}
\setlength{\tabcolsep}{4pt}
\begin{tabular}{clllrrrrr}
\toprule
\textsc{\#} & \textsc{model} & \textsc{type} & \textsc{arch} & \textsc{wer}$\downarrow$ & \textsc{sub} & \textsc{ins} & \textsc{del} & \textsc{cer}$\downarrow$ \\
\midrule
\emph{ref} & \emph{ABAIR / Fotheidil} & \emph{closed} & \emph{TDNN-HMM} & \emph{44.5*} & --- & --- & --- & --- \\
\midrule
1 & omniASR LLM 7B & open-w & w2v2+Trans & \textbf{39.1} & 32.2 & 3.4 & 3.5 & \textbf{18.6} \\
2 & omniASR LLM 300M & open-w & w2v2+Trans & 47.7 & 38.4 & 4.9 & 4.4 & 24.1 \\
3 & azure / speech-ga-IE & API & proprietary & 57.5 & 21.5 & 3.5 & 32.5 & 43.8 \\
4 & mms-1b-all & open & w2v2 CTC & 61.6 & 51.9 & 3.2 & 6.5 & 26.0 \\
5 & Aditya3107 / xls-r-1b & open & w2v2 CTC & 75.8 & 62.1 & 6.2 & 7.5 & 36.0 \\
6 & kingabzpro / xls-r-1b & open & w2v2 CTC & 78.5 & 64.7 & 9.2 & 4.6 & 38.2 \\
7 & jimregan / xlsr-53 & open & w2v2 CTC & 83.0 & 68.2 & 9.8 & 5.0 & 40.8 \\
8 & cpierse / xlsr-53 & open & w2v2 CTC & 83.2 & 68.6 & 9.8 & 4.8 & 41.9 \\
\midrule
9 & whisper-large-v2 & open & enc-dec & 102.8 & 78.2 & 19.8 & 4.8 & 59.6 \\
10 & whisper-medium & open & enc-dec & 134.1 & 86.5 & 43.1 & 4.4 & 87.3 \\
11 & whisper-large-v3 & open & enc-dec & 217.8 & 89.8 & 123.7 & 4.3 & 156.0 \\
12 & whisper-large-v3-turbo & open & enc-dec & 587.6 & 91.2 & 491.2 & 5.1 & 410.1 \\
\bottomrule
\end{tabular}
\end{table*}

\section{Benchmark Findings}

\subsection{The Whisper variants we evaluated fail catastrophically on Irish}

All four Whisper variants exceed 100\%~WER on both datasets, with insertion rates of 20--491\%: the decoder emits fluent English unrelated to the input (Appendix~\ref{app:whisper-outputs}). v3 is worse than v2 (125.6\% vs 106.0\% on CV), and turbo is worse again (587.6\% on FLEURS). \citet{qian2024whisper} report 110.4\%~WER for whisper-large-v3 on FLEURS Irish, which is in the same regime as our results; we did not find prior documentation of the monotonic v2~$\to$~v3~$\to$~turbo regression on Irish.

\subsection{Common Voice overestimates performance}
\label{sec:generalization}

\begin{table}[ht]
\centering
\small
\caption{Cross-corpus generalisation gap. $\Delta$ = FLEURS $-$ CV WER. Multilingual models generalise; CV-trained models do not.}
\label{tab:gap}
\begin{tabular}{lrrr}
\toprule
\textsc{model} & \textsc{cv} & \textsc{fleurs} & $\Delta$ \\
\midrule
mms-1b-all & 54.2 & 61.6 & \textbf{+7.3} \\
omniASR 7B & 30.6 & 39.1 & +8.4 \\
omniASR 300M & 37.6 & 47.7 & +10.1 \\
\midrule
azure & 22.2 & 57.5 & +35.2 \\
cpierse & 49.4 & 83.2 & +33.8 \\
Aditya3107 & 32.4 & 75.8 & \textbf{+43.4} \\
\bottomrule
\end{tabular}
\end{table}

Models fine-tuned on Common Voice degrade 33--43 WER points on FLEURS (Table~\ref{tab:gap}). Azure degrades 35 points. Models with massively multilingual pre-training degrade only 7--10 points. Prior work has evaluated across corpora \citep{lonergan2025fotheidil}, but the magnitude of this gap has not been quantified under a single harness. Common Voice WER alone is not a reliable proxy for deployment; BlasBench therefore requires evaluation on at least two datasets.

\subsection{The open--closed gap}

ABAIR's Fotheidil reports its best results (10.9\%~WER on private M\'{i}leGl\'{o}r, 19.6\% on Common Voice, and 44.5\% on FLEURS-R) using their full pipeline including RNNLM rescoring; without LM rescoring the M1 acoustic model alone reports 23.7\% on Common Voice \citep{lonergan2025fotheidil, ma2024fleursr}.\footnote{ABAIR evaluates on FLEURS-R (audio restored from the original FLEURS via the Miipher model), not the original FLEURS we use, and with a different normaliser. Numbers are indicative, not directly comparable.} omniASR~7B reaches 30.6\% and 39.1\% on original FLEURS under our normaliser. Because datasets and normalisers differ, we treat these numbers as indicative rather than a direct open-vs-closed measurement; they are consistent with a narrower gap than the headline 10.9\%/30.6\% suggests, but do not establish one. ABAIR's architecture (TDNN-HMM, Kaldi-based) is from a different family than the strongest open models here (wav2vec2-based), and their reported gains coincide with substantial Irish-specific training data (398h labelled, 3{,}230h pseudo-labelled \citep{lonergan2025fotheidil}) that exceeds what any open model was trained on for Irish. The primary binding constraint on Irish ASR is data, not architecture.

\section{Related Work}

\paragraph{Irish ASR.} ABAIR has published on hybrid systems \citep{lonergan2022abair}, dialect lexicons \citep{lonergan2022lexicon}, dialect-balanced training \citep{lonergan2023balanced}, dialect identification \citep{lonergan2023dialect}, multi-task learning \citep{lonergan2024multitask}, and Fotheidil \citep{lonergan2025fotheidil}. \citet{qian2024whisper} fine-tune Whisper for Irish but compare only Whisper variants. To our knowledge, no prior work compares multiple end-user Irish ASR systems from different architecture families under a single Irish-aware evaluation protocol.

\paragraph{Low-resource and Celtic-language benchmarks.} \citet{arif2024wer} benchmark Whisper, MMS, and SeamlessM4T on Urdu with corpus-level WER aggregation; BlasBench adopts a similar aggregation strategy and adds bootstrap CIs and cross-corpus analysis. \citet{klejch2025gaelic} build Scottish Gaelic ASR with Gaelic-specific normalisation, \citet{jones2022welsh} fine-tune wav2vec2 for Welsh, and \citet{bartley2025manxcornish} demonstrate minimal-data ASR for Manx and Cornish from a spoken dictionary. On evaluation methodology, \citet{manohar2024normalization} show that multilingual ASR normalisers can distort error rates when they erase language-specific orthographic information. On statistical methodology, \citet{gillick1989statistical} and \citet{bisani2004bootstrap} remain the canonical references for significance testing of WER comparisons.

\paragraph{Multilingual foundation models and benchmarks.} Beyond MMS and Whisper, \citet{zhang2023usm} (Google USM), \citet{puvvada2024canary} (NVIDIA Canary), and \citet{peng2024owsmctc} (OWSM-CTC) all target broad multilingual speech coverage, but none of these papers establishes end-user Irish ASR performance. Multilingual benchmarks FLEURS \citep{conneau2022fleurs}, ML-SUPERB \citep{shi2023mlsuperb}, and XTREME-S \citep{conneau2022xtremes} include Irish among many languages but apply no Irish-specific normalisation; the Open ASR Leaderboard \citep{srivastav2025openasrlb} does not include Irish at all (its multilingual track currently covers German, French, Italian, Spanish, and Portuguese). \citet{barry2022gabert} release gaBERT for Irish NLP but not for ASR evaluation.

\section{Conclusion}

BlasBench is, to our knowledge, the first open Irish-specific benchmark that compares end-user ASR systems under a shared Irish-aware protocol. Across 12~systems on two datasets, wav2vec2 models are the only viable family we tested, the Whisper variants we evaluate fail catastrophically, and the gap between Common Voice and FLEURS scores exposes a generalisation failure that single-dataset evaluation hides. The deeper bottleneck is data: labelled Irish audio remains the binding constraint on progress. Because scoring is deterministic and reproducible, BlasBench is suitable as an evaluation layer within automated research pipelines \citep{karpathy2026}, where reliable comparison across many candidate models depends on a stable scoring function. The harness, normaliser, per-utterance predictions, and run metadata are publicly released.

\section*{Limitations}

We evaluate on fixed public BlasBench slices for Common Voice ga-IE (874 utterances) and FLEURS ga-IE (842 utterances); both consist of read speech, and no public conversational Irish ASR benchmark exists. These sample sizes limit the precision of pairwise comparisons. Confidence intervals are reported in the accompanying JSON; main tables show point estimates for compactness, as noted in the evaluation pipeline.

Neither dataset spans all three dialects equally, and neither provides dialect labels, reflecting a gap in available Irish data rather than a design choice. Dialect-stratified evaluation is not possible with current public resources.

The 12 systems evaluated cover the main open, community, and commercial categories; selection was constrained by licence, cost, and available credentials. OmniASR was run outside the shared evaluation harness because a fairseq2 driver dependency was unresolved at the time of writing, with a harness adapter pending; its numbers are therefore not directly comparable at the implementation level. Azure results depend on the API version and model snapshot available at time of evaluation (April 2026); future versions may differ.

ABAIR uses FLEURS-R rather than original FLEURS as its test set, and its normaliser is not publicly available; the open--closed comparison is therefore indicative rather than controlled. The WER impact of our normalisation choices has not been quantified.

The community wav2vec2 models were fine-tuned on earlier Common Voice releases (CV~6.1--8.0) and evaluated on CV~25.0. Some of the cross-corpus gap likely reflects distribution shift alongside genuine generalisation behaviour, with the multilingual-pretrained models in Table~\ref{tab:gap} providing a partial control.


\section*{Acknowledgements}
Portions of this work were assisted by Claude Opus 4.6 \citep{claude2026}.

\clearpage
\appendix

\section{Error-Type Breakdown}
\label{app:error-types}

WER alone conflates three failure modes that point at different fixes. Table~\ref{tab:error-breakdown} reports S, I, and D as percentages of total reference words, aggregated globally.

\begin{table}[ht]
\centering
\small
\setlength{\tabcolsep}{6pt}
\renewcommand{\arraystretch}{1.05}
\caption{Error-type breakdown. S, I, D reported as percentages of reference words, aggregated globally. Sorted by WER within each dataset.}
\label{tab:error-breakdown}
\begin{tabular}{lrrrr}
\toprule
\textsc{model} & \textsc{wer} & \textsc{s} & \textsc{i} & \textsc{d} \\
\midrule
\multicolumn{5}{l}{\textit{Common Voice}} \\
\midrule
azure ga-IE              & 22.3  & 15.8 & 1.7   & 4.8  \\
omniASR 7B               & 30.7  & 25.0 & 2.5   & 3.2  \\
Aditya3107 xls-r-1b      & 32.4  & 26.4 & 1.7   & 4.3  \\
omniASR 300M             & 37.6  & 29.3 & 3.4   & 4.9  \\
kingabzpro xls-r-1b      & 45.8  & 38.2 & 3.6   & 4.0  \\
jimregan xlsr-53         & 48.9  & 40.5 & 4.2   & 4.2  \\
cpierse xlsr-53          & 49.4  & 41.5 & 3.9   & 4.0  \\
mms-1b-all               & 54.3  & 44.1 & 2.8   & 7.4  \\
whisper-large-v2         & 106.0 & 73.6 & 20.0  & 12.4 \\
whisper-large-v3         & 125.6 & 78.8 & 33.1  & 13.7 \\
whisper-medium           & 129.3 & 76.2 & 40.6  & 12.5 \\
whisper-large-v3-turbo   & 225.6 & 83.1 & 128.8 & 13.7 \\
\midrule
\multicolumn{5}{l}{\textit{FLEURS}} \\
\midrule
omniASR 7B               & 39.1  & 32.2 & 3.4   & 3.5  \\
omniASR 300M             & 47.7  & 38.4 & 4.9   & 4.4  \\
azure ga-IE              & 57.5  & 21.5 & 3.5   & 32.5 \\
mms-1b-all               & 61.6  & 51.9 & 3.2   & 6.5  \\
Aditya3107 xls-r-1b      & 75.8  & 62.1 & 6.2   & 7.5  \\
kingabzpro xls-r-1b      & 78.5  & 64.7 & 9.2   & 4.6  \\
jimregan xlsr-53         & 83.0  & 68.2 & 9.8   & 5.0  \\
cpierse xlsr-53          & 83.2  & 68.6 & 9.8   & 4.8  \\
whisper-large-v2         & 102.8 & 78.2 & 19.8  & 4.8  \\
whisper-medium           & 134.1 & 86.5 & 43.1  & 4.4  \\
whisper-large-v3         & 217.8 & 89.8 & 123.7 & 4.3  \\
whisper-large-v3-turbo   & 587.6 & 91.2 & 491.2 & 5.1  \\
\bottomrule
\end{tabular}
\end{table}

wav2vec2 errors are substitutions; Whisper errors are insertions. Every wav2vec2 model keeps I below 10\%; every Whisper variant exceeds 20\%, peaking at 491\% for large-v3-turbo on FLEURS. That is not an acoustic failure (see Appendix~\ref{app:whisper-outputs}). Going CV~$\to$~FLEURS, wav2vec2 S and I both roughly double while D stays flat; Whisper I grows another order of magnitude---longer audio gives the decoder more runway to hallucinate. Azure is the anomaly: S-dominated on CV, D-dominated on FLEURS (4.8~$\to$~32.5), suggesting a VAD or endpointer tuned for short clips.

\section{What the Hard Utterances Look Like}
\label{app:hard-utterances}

Filtering to utterances where every non-Whisper model exceeds 50\%~WER isolates text-level difficulty. Two distinct tails emerge.

\textbf{Common Voice} hard cases are short (4--6 words, where one error is 15--25~WER points by construction) and Irish-internal: dialect forms (\emph{chuile}, \emph{bréidín}, \emph{comhluadar}, \emph{airneáin}), copula and conditional idioms (\emph{gréasaí ab ea é}, \emph{ní chuirfinn thairis é}), and Gaeltacht place names in their eclipsed form (\emph{i~ngaillimh~thiar}, \emph{leitir~mealláin}). Hard because training data underrepresents them, not because the text is exotic.

\textbf{FLEURS} hard cases are code-switched news text: alphanumerics fused into single tokens (\emph{80211n}, \emph{24ghz}, \emph{m16}, \emph{10001100}), foreign proper nouns (\emph{wong kan seng}, \emph{rolanda mendoza}, \emph{aerosmith}), English-origin acronyms pronounced as English (\emph{mdt}, \emph{swapo}), and bare in-text academic citations (\emph{larson agus lafasto 1989 lch 109}). A number-and-acronym expansion rule would partially fix this tail.

\section{Representative Whisper Outputs}
\label{app:whisper-outputs}

Four examples showing what Whisper's insertions actually are. \textsc{ref}~=~Irish reference, \textsc{w3}~=~whisper-large-v3, \textsc{wv2}~=~\texttt{Aditya3107/xls-r-1b} (acoustically weak, does not hallucinate).

\begin{quote}\small
\textit{(a) Whisper emits Welsh} \quad \textsc{cv} 545\\
\textsc{ref}\ \ \ dia dhaoibh tráthnóna\\
\textsc{w3}\ \ \ \ diolch yn fawr iawn am wylior fideo\\
\textsc{wv2}\ \ dia dhaoibh tráthnóna
\end{quote}

\begin{quote}\small
\textit{(b) Repetition loop} \quad \textsc{cv} 216 (turbo)\\
\textsc{ref}\ \ \ tabhair cabhair don fhoireann\\
\textsc{w3}\ \ \ \ to a coward to a coward to a coward \ldots (333 tokens)\\
\textsc{wv2}\ \ tabhair cabhair don fhoireann
\end{quote}

\begin{quote}\small
\textit{(c) Unrelated English sign-off} \quad \textsc{fl} 609\\
\textsc{ref}\ \ \ phléasc buama amháin lasmuigh doifig an ardghobharnóra\\
\textsc{w3}\ \ \ \ thank you for listening and have a good day\\
\textsc{wv2}\ \ pléis buam amhain leasmúid duifigh an ard gabhrana
\end{quote}

\begin{quote}\small
\textit{(d) Topic-adjacent English} \quad \textsc{fl} 98 (turbo)\\
\textsc{ref}\ \ \ ina dhiaidh sin bogadh chuig ospidéal addenbrooke i gcambridge é\\
\textsc{w3}\ \ \ \ in the next day ill be back to edinburghs hospital in cambridge\\
\textsc{wv2}\ \ ina dhíg sin bothar chuig ospadéal adan bhrog a ceamraid é
\end{quote}

Case~(d) is the telling one: Whisper catches \emph{cambridge}, hears \emph{addenbrooke} as \emph{edinburgh}, and wraps them in fluent English syntax. The acoustic signal reaches the decoder; the decoder refuses to emit Irish. Irish is a rounding error in Whisper's token budget (\S\ref{sec:generalization}), so its strongest prior on any audio segment is a language it knows.

\end{document}